\documentclass[preprints,article,accept,moreauthors,pdftex]{Definitions/mdpi} 

\firstpage{1} 
\pubvolume{1}
\issuenum{1}
\articlenumber{0}
\pubyear{2021}
\copyrightyear{2020}
\datereceived{} 
\dateaccepted{} 
\datepublished{} 
\hreflink{https://doi.org/} 

\Title{Visual diagnosis of the Varroa destructor parasitic mite in honeybees using object detector techniques}

\TitleCitation{Visual diagnosis of the Varroa destructor parasitic mite in honeybees using object detector techniques}

\Author{Simon Bilik $^{1}$\orcidA{},
        Lukas Kratochvila $^{2}$\orcidB{},
        Adam Ligocki $^{3}$\orcidC{},
        Ondrej Bostik $^{4}$\orcidD{},
        Tomas Zemcik $^{5}$\orcidE{},
        Matous Hybl $^{6}$\orcidF{},
        Karel Horak $^{7}$\orcidG{} and
        Ludek Zalud $^{8}$\orcidH{}}

\AuthorNames{Simon Bilik, 
             Lukas Kratochvila,
             Adam Ligocki,
             Ondrej Bostik,
             Tomas Zemcik,
             Matous Hybl,
             Karel Horak,
             Ludek Zalud}

\AuthorCitation{Bilik, S.; Kratochvila, L.; Ligocki, A.; Bostik, O.; Zemcik, T.; Hybl, M.; Horak, K.; Zalud, L.}

\address{%
$^{1}$ \quad bilik@feec.vutbr.cz\\
$^{2}$ \quad kratochvila@feec.vutbr.cz\\
$^{3}$ \quad adam.ligocki@vutbr.cz\\
$^{4}$ \quad bostik@feec.vutbr.cz\\
$^{5}$ \quad zemcikt@feec.vutbr.cz\\
$^{6}$ \quad matous.hybl@ceitec.vutbr.cz\\
$^{7}$ \quad horak@feec.vutbr.cz\\
$^{8}$ \quad zalud@feec.vutbr.cz}

\corres{Correspondence: bilik@feec.vutbr.cz}

\abstract{The \textit{Varroa destructor} mite is one of the most dangerous Honey Bee (\textit{Apis mellifera}) parasites worldwide and the bee colonies have to be regularly monitored in order to control its spread. Here we present an object detector based method for health state monitoring of bee colonies. This method has the potential for online measurement and processing. In our experiment, we compare the YOLO and SSD object detectors along with the Deep SVDD anomaly detector. Based on the custom dataset with 600 ground-truth images of healthy and infected bees in various scenes, the detectors reached the highest F1 score up to 0.874 in the infected bee detection and up to 0.727 in the detection of the \textit{Varroa destructor} mite itself. The results demonstrate the potential of this approach, which will be later used in the real-time computer vision based honey bee inspection system. To the best of our knowledge, this study is the first one using object detectors for this purpose. We expect that performance of those object detectors will enable us to inspect the health status of the honey bee colonies in real time.}


\begin{document}

\section{Introduction}

\textit{Varroa destructor mite} (V.-mite) is a honey bee (\textit{Apis mellifera}) ectoparasite, which is spread around the whole world, except Australia \cite{rosenkranz2010biology}. This mite attacks the bees in the larval state and it is a serious threat to bee colonies. Infected bees are often deformed due to the development defects and the surviving bees are weakened as the V.-mite feeds on them \cite{ramsey2019varroa}. It has also proven that the V.-mite transmits several viruses. All those factors can cause serious, even fatal bee colony losses. This parasite is also connected with the Colony Collapse Disorder, which caused huge losses in several states worldwide \cite{genersch2010german}.

\begin{figure}[H]
    \centering
    \includegraphics[width=8.6cm]{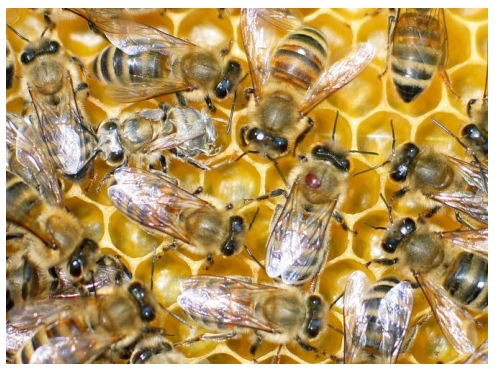}
    \caption{V.-mite on an adult bee worker (in the centre)  \cite{rosenkranz2010biology} Reprinted from Publication Journal of Invertebrate Pathology, 103, P. Rosenkranz; P. Aumeier; B. Ziegelmann, Biology and control of Varroa destructor, 96-101, Copyright (2010), with permission from Elsevier}
    \label{fig:bees}
\end{figure}

The V.-mite originally descended from East Asia and it started to spread around the world around the half of the 20th century, the western honey bee is not adapted for this parasite. Due to this factor, periodic V.-mite monitoring and treatment has to be performed. The conventional monitoring methods include the inspection of the bee colony detritus, or the V.-mite isolation from the bee sample, but those methods are time consuming and labour intensive. The V.-mite also becomes resistant to the current medications and new treatment methods, or approaches have to be developed \cite{rosenkranz2010biology}.

In this paper we present a non-invasive computer vision (CV) based approach for the V.-mite monitoring, which can be later used as a foundation of a portable online monitoring system.

Our paper is structured as follows: In the first part, we bring a short overview of the existing bee monitoring methods with the emphasis on the existing V.-mite automatic detection system and the vision based honey bee monitoring systems. The second part describes the techniques used in our experiment and the dataset we used. In the end, we discuss the experiment’s results and we provide suggestions for the further research.

\section{Related Work}

Several studies focus on various honey bee monitoring methods, for example the sound analysis for the swarm detection in \cite{ferrari2008monitoring}, bee activity monitoring in \cite{mezquida2009platform}, or the RFID based individual bee tracking in \cite{de2018low}. A brief overview of other and above mentioned honeybee monitoring techniques can be also found in \cite{BUT164026}.

Of the image processing based methods, the paper \cite{giuffre2017automated} focuses on bee autogrooming phenomena. This study develops a new methodology of modeling the autogrooming process. The bees are covered by baking flour and their grooming is observed by a camera on a controlled background. The speed of the grooming process is automatically analysed. 

In \cite{knauer2007comparison} the authors try to classify the honeybee brood cells into three categories: occluded, visible and closed, and visible and uncapped for the bee colony self-hygiene monitoring. Authors develop a simple system for the honeycomb’s optical inspection, where they compare the performance of several classifiers. The evaluated classifying algorithms are: support vector machine (SVM), decision tree and boosted classifiers. The best results are obtained with the SVM.

The paper \cite{rodriguez2018recognition} presents a CNN based approach to bee classification based on whether or not the bees are carrying pollen. Hardware setup for image acquisition is described, methods for bee segmentation are discussed and VGG16, VGG19 and ResNet50 CNN performance is compared to classical classifiers such as KNN, Naive Bayes and SVM. Rodriguez et al. conclude that shallower models such as VGG have better performance compared to deeper models such as ResNet. Generally performance of CNN based classifiers was superior to conventional classifiers. What is also important is the discovery that the use of a human predefined colour model, needed for classical classifiers, influenced only the training time but not the final performance of the CNNs when compared to straight RGB input. The authors also created a publicly available dataset \cite{rodriguez_dataset} with high-resolution images of the pollen bearing and non-pollen bearing bees with 710 image samples in total.

The investigated V.-mite detection techniques are described in the following articles: The study presented in \cite{szczurek2019detection} is focused on detection of V.-mite infestation of   honeybee colonies. This study is based on beehive air measurements using an array of partially selective gas sensors in field condition. Collected data were used for training simple LDA and k-NN classifiers with the best misclassification error for k-NN were 17\%. The results indicated a possibility to detect honey bee diseases with this approach, but the method must be improved before deployment. In particular, the big and expensive data collection hardware alongside the classifier need to be improved.

The paper \cite{bauer2018recognition} shows that the V.-mite-infected brood cells are slightly warmer than the uninfested ones. The experiment was based on the thermal video sensing of brood frames with the artificially infected and uninfected cells in the controlled thermal conditions. The authors observed that infected cells are slightly warmer (between 0.03 and 0.19°C), which shows the possibility of the thermal measurements in order to detect mite-infected broods.

The article presented in \cite{elizondo2013video} shows a method of monitoring the V.-mite’s motion on the bee brood samples. The authors use classical computer vision methods as background subtraction and geometric pattern analysis with double thresholding to detect the mites on the video frames followed by their tracking on the bee brood with the accuracy rate of 91\%. Anyway, this method is not suitable for in-field-analysis, because the brood must be removed from the cell before the analysis.

The study \cite{schurischuster2016sensor} describes the experimental setup for a non-destructive V.-mite detection and brings suggestions for further development. It also identifies the challenges for mite detection, for example as motion blur, mite colour, or reflections. This paper only briefly describes possible segmentation and detection techniques.

The following article \cite{schurischuster2018preliminary} from the same authors describes bee detection and their classification, using classical computer vision methods. In the first part, the authors present good results in segmentation of individual bees using a combination of the Gaussian Mixture Models and the color thresholding based methods. In the second part, they provide the results of a two class classification (healthy bee/bee with mite) using the Naive Bayes, SVM and the Random forest classifiers. The accuracy of those methods varies according to the frame preprocessing techniques and in the best cases moves around 80\%.

The last article \cite{schurischuster2020image} of those authors extends the article \cite{schurischuster2018preliminary} with the CNN based classificators (ResNet and AlexNet) and the DeepLabV3 semantic segmentation. The processing pipeline and the classification classes remain the same, as it was in the previous article. Classification of images with a single bee shows good results with the accuracy for the “infected” class around 86\%. The experiments with the bees classification from the whole image using the sliding window bring weaker results with the false-negative rate around 67

The study presented in \cite{bjerge2019computer} builds on the \cite{schurischuster2016sensor}, \cite{schurischuster2018preliminary} and \cite{rodriguez2018recognition}. The authors present there a whole V.-mite detection pipeline. Firstly, they describe a video monitoring unit with the multispectral illumination, which could be connected directly to the beehive through several narrow pathways. The captured frames were recorded and processed offline, as the bees were segmented with classical computer vision methods and the mites were detected with a custom CNN model. The authors achieved good results in the mite infestation estimation and they proved the potential of the CV based on-site testing.

Practical applications stemming from efficient bee monitoring systems and V.-mite detection systems have also been studied. The authors of \cite{Laser7850001} investigate the possibility of using a camera based bee monitoring system on the hive’s entrance to detect V.-mites and then destroy them with a focused laser beam. The authors outlined hardware and software requirements for such a system and found it feasible even if such systems have not yet been developed to a deployable state.

\section{Materials and Methods}

The goal of our experiment was to prove whether the state-of-the-art object detectors YOLOv5 and SSD can perform the V. mite and bee detection, alternatively to detect and distinguish between the healthy and the infected bee. Unlike the \cite{bjerge2019computer}, \cite{schurischuster2018preliminary} and \cite{schurischuster2020image}, we don’t separate object segmentation and classification, which could lead to faster results and to the possibility of online processing. 

In this chapter, we firstly describe our dataset and its statistics. Then follows a brief description of the used detector’s architectures along with the hyperparameters and the evaluation metrics.

\subsection{Dataset Description}

During the initial phase of our research, we found several publicly available honey-bee datasets. The dataset from \cite{rodriguez2018recognition} is designed for the pollen-wearing bee recognition only and despite its quality and high resolution, it was unsuitable for our task. The dataset presented in \cite{schurischuster2020image} contains short videos with mite-infected and uninfected bees in high resolution and it was partially used in our experiment. The dataset \cite{kaggle} seemed promising and brought a lot of additional information, but the images were in a bad quality and low resolution.

\begin{figure}[H]
    \centering
    \includegraphics[width=12cm]{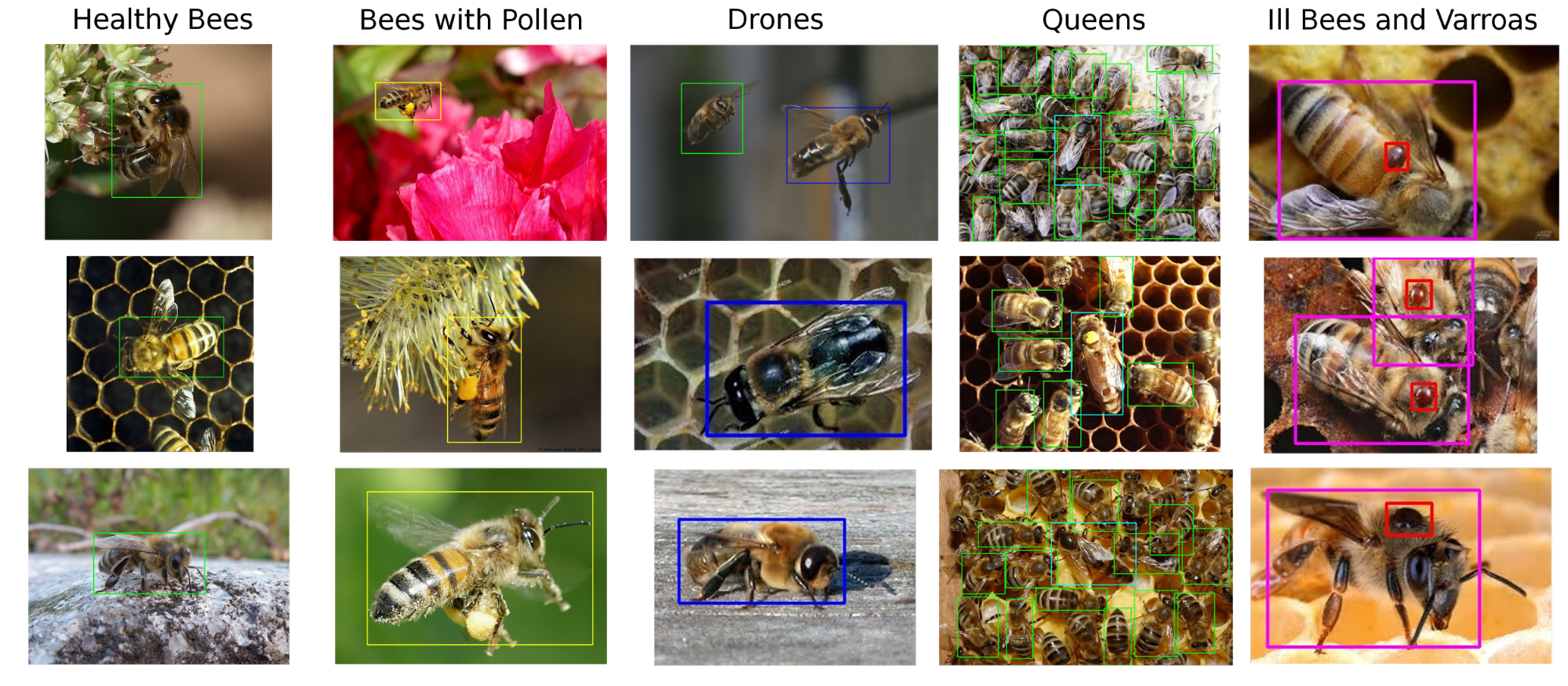}
    \caption{Brief overview of dataset created for the purpose of this work. Healthy bees (green), bees with pollen (yellow), drones (blue), queens (cyan), infected bees (purple), V.-mite (red).}
    \label{fig:bees_classes}
\end{figure}

For the above reasons, our custom dataset was designed from publicly available images and partially from the dataset \cite{schurischuster_stefan_2020_4085043}. It contains a total of 803 unique  samples, where 500 samples capture bees in the general environment and the rest, taken from the dataset \cite{schurischuster_stefan_2020_4085043}, shows bees on an artificial background.

\begin{figure}[H]	
\centering
\widefigure
    \includegraphics[width=15cm]{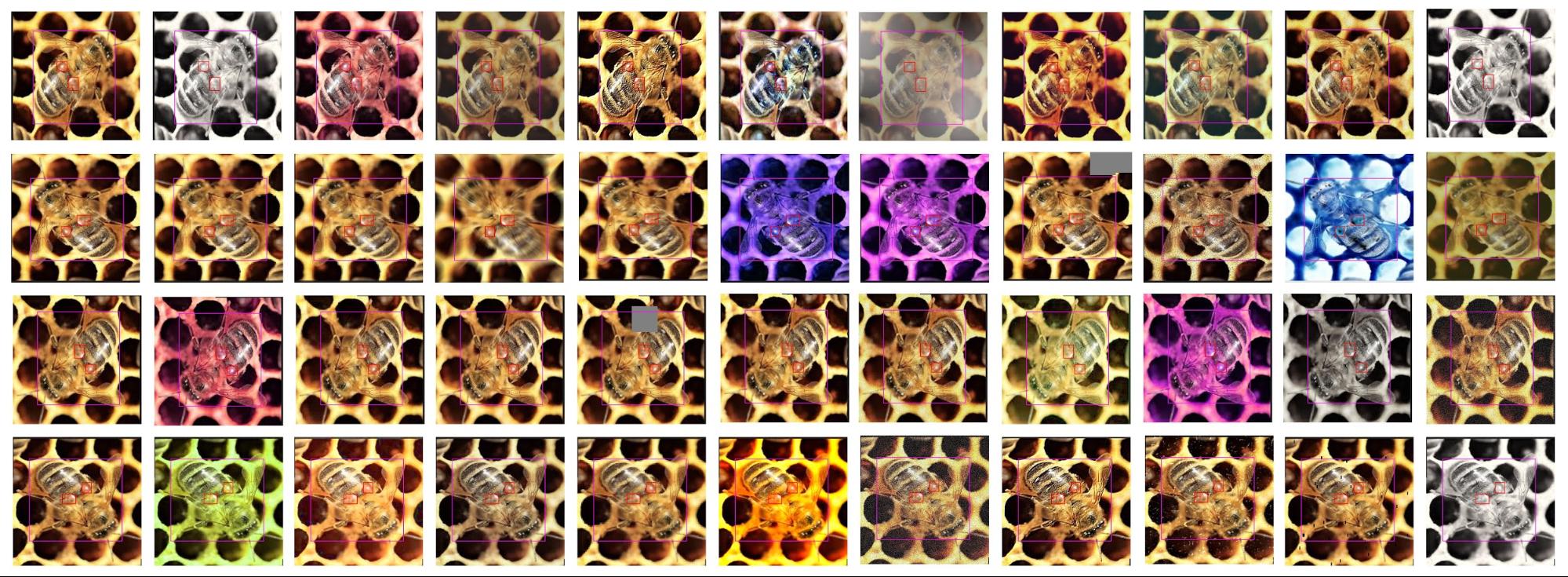}
    \caption{Example of a single image (left top) augmented into the ten more training samples. In the top left column, there is the original image rotated by 0, 90, 180, and 270 degrees. In the rest of the row, there are the images augmented by the ImgAug framework.}
    \label{fig:bees_augs}
\end{figure}

In our dataset and with future experimentation in mind, we define six classes shown in figure \ref{fig:bees_augs}, the healthy bee (1), the bee with pollen (2) (pollen may be miss-classified with a V.-mite), the drone (3), the queen (4), V.-mite-infected bee (5), and the V.-mite (6). However, only for this paper, we reduced the data annotation into three different subsets. The first one is the bees (classes 1, 2, 3, 4, 5) and V.-mite (class 6), the second one is the healthy bees (classes 1, 2, 3, 4) and infected bees (class 5), finally in the last subset are the data annotated only with V.-mite (only class 6). 

We created these three derivations of the original dataset annotation to test, which one will give us the best result in detecting varroosis infection in the beehive. Our dataset was manually annotated with the LabelImg tool \cite{labelImg}, and its statistics are presented in table \ref{tab:dataset}. All annotations were consulted with a professional beekeeper.

Even though we used a part of the images from the dataset \cite{schurischuster_stefan_2020_4085043}, the number of gathered images to train the neural network was not sufficient. Therefore we have decided to apply an augmentation to the entire training subset. Generally speaking, augmentation helps increase the robustness of the training process, helps to avoid overfitting, and overall improves the neural network's performance at the end of the training phase \cite{shorten2019survey}.

To apply augmentation to our data, we have used the Python library called ImgAug \cite{imgaug}. It provides a wide range of various methods to augment the image by blurring, adding different noise signals, various effects (like motion, fog, rain, color space alteration, etc.), random erasing (Random Erasing Data Augmentation) bounding box modifications, or even the color space modifications and shifting or image cutouts \cite{shorten2019survey}, \cite{zoph2020learning}.

The main idea of the augmentation method is to create slightly modified derivatives of the original training data that, even after modification, still represents the original problem. In the case of object detection or, generally speaking, computer vision, by applying rotation, geometrical distortion, additional noise, or the colour shift, we do not modify the information that the image contains. Later, during the training process, the neural network is forced to learn how to solve the given task for perfect-looking training data as well as, for distorted, noise-added, or colour shifted images. It makes models to better generalize problems, and the entire learning process is way more robust against overfitting.

In the case of V.-mite detection, we can illustrate the problem of the close similarity between the V.-mite and the bee's eye. Both objects are very close to each other in their geometrical shapes and dimensions. The differences here are the colour and the close surroundings. The mite is strictly brown whereas the bee's eyes are black. If we let the neural network to train on the unaugmented images, it could learn to identify mites only by the presence of the brown colour. If the images are augmented, the neural network has to understand the entire structure of the mite body.

\end{paracol}
\begin{paracol}{2}
\switchcolumn

In our case, we rotated each image by 90, 180, and 270 degrees, and for every rotation, we have created ten derivatives of images by applying a single, randomly selected augmentation style from the available set. Using this method, we had augmented every single image from the original training set and we created additional 43 samples. It is important to note that the validation and the test set stay without any modifications, so they represent the bees and V.-mites real-life image data.

In total, we created the dataset containing 803 images. The training set contains 561 images, later augmented for the total of 24684 images, 127 images comprise the validation set, and the test set contains further 115 samples. All images in the original 803-images set are independent of each other. For those 803 images, we created three annotations as mentioned before, and the number of annotated instances in each class for training, validation and test set is in table \ref{tab:dataset}.

\clearpage

\end{paracol}
\begin{specialtable}[H]
\scriptsize
\centering

\begin{tabular}{|c|c|c|c|c|c|c|c|c|}
\hline
\multicolumn{2}{|c|}{\multirow{3}{*}{\begin{tabular}[c]{@{}c@{}}All Classes\\ Dataset\end{tabular}}} & \multicolumn{6}{c|}{No. of Annotated Objects per Clas} & \multirow{2}{*}{Images} \\ \cline{3-8}
\multicolumn{2}{|c|}{} & \begin{tabular}[c]{@{}c@{}}Bee Worker\\ (No Pollen)\end{tabular} & \begin{tabular}[c]{@{}c@{}}Bee Worker \\ (Pollen)\end{tabular} & \begin{tabular}[c]{@{}c@{}}Bee \\ Drone\end{tabular} & \begin{tabular}[c]{@{}c@{}}Bee \\ Queen\end{tabular} & \begin{tabular}[c]{@{}c@{}}Bee With \\ V.mites\end{tabular} & V.- Mite &  \\ \cline{3-9} 
\multicolumn{2}{|c|}{} & 1158 & 143 & 19 & 52 & 298 & 424 & 803 \\ \hline\hline
\multirow{5}{*}{\begin{tabular}[c]{@{}c@{}}Bees and \\ V.-mites \\ Dataset\end{tabular}} & Classes & \multicolumn{5}{c|}{Bees} & V.-Mite & - \\ \cline{2-9} 
 & Train Set & \multicolumn{5}{c|}{1148} & 250 & 561 \\ \cline{2-9} 
 & Train Aug Set & \multicolumn{5}{c|}{50512} & 11000 & 24684 \\ \cline{2-9} 
 & Val Set & \multicolumn{5}{c|}{274} & 92 & 127 \\ \cline{2-9} 
 & Test Set & \multicolumn{5}{c|}{248} & 59 & 115 \\ \hline\hline
\multirow{5}{*}{\begin{tabular}[c]{@{}c@{}}Healthy and Ill \\ Bees Dataset\end{tabular}} & Classes & \multicolumn{4}{c|}{Healthy Bees} & Infected Bees & - & - \\ \cline{2-9} 
 & Train Set & \multicolumn{4}{c|}{956} & 192 & - & 561 \\ \cline{2-9} 
 & Train Aug Set & \multicolumn{4}{c|}{42064} & 8448 & - & 24684 \\ \cline{2-9} 
 & Val Set & \multicolumn{4}{c|}{220} & 54 & - & 127 \\ \cline{2-9} 
 & Test Set & \multicolumn{4}{c|}{196} & 52 & - & 115 \\ \hline\hline
\multirow{5}{*}{\begin{tabular}[c]{@{}c@{}}V.-Mites \\ Dataset-\end{tabular}} & Classes & \multicolumn{5}{c|}{-} & V.-Mite & - \\ \cline{2-9} 
 & Train Set & \multicolumn{5}{c|}{-} & 250 & 561 \\ \cline{2-9} 
 & Train Aug Set & \multicolumn{5}{c|}{-} & 11000 & 24684 \\ \cline{2-9} 
 & Val Set & \multicolumn{5}{c|}{-} & 92 & 127 \\ \cline{2-9} 
 & Test Set & \multicolumn{5}{c|}{-} & 59 & 115 \\ \hline

\end{tabular}

\caption{The table shows the number of annotated objects of given classes in three versions of the dataset created by reducing the number of classes in different ways (bees and varroosis, healthy and infected bees and the varroosis only dataset).
}
\label{tab:dataset}
\end{specialtable}
\begin{paracol}{2}
\switchcolumn

We are aware that for image classification tasks it is crucial to have a balanced dataset for all classes. In case a balanced dataset is not available, the architecture will not be able to train classifying the underrepresented classes. In order to identify this potential problem in our slightly unbalanced dataset, we use the mAP[0.5] score as one of our metrics. This score is affected by all classes with the same weight and it is described in more detail below.

\subsection{Network Description}

In this section we provide a brief description of the object detectors YOLOv5 and SSD along with the Deep SVDD anomaly detector used in our experiment.

\subsubsection{YOLOv5}

The original YOLO \cite{redmon2016you} and all derived neural networks (YOLO9000, YOLOv3, YOLOv4) are examples of end-to-end object detection models. It means the inference of the network with the image is the only operation performed during the object detection. There is nothing like region propositioning, combining detection's bounding boxes, and so on \cite{ren2016faster}. The YOLO architecture has only the image on the input and the vector of detections and probabilities of these detections on the output.

In this paper, we have used the open-source implementation of the YOLO object detector, called YOLOv5 from Ultralytics, available from \cite{jocher2020yolov5}. The YOLOv5 is not a self-standing version, that would bring significant improvements in the YOLO-like neural networks architectures, more it is one example of implementing the YOLOv3 principles in the PyTorch framework \cite{NEURIPS2019_9015}. Also, as of the day of writing this paper, there is no official peer-reviewed article about YOLOv5.

The YOLOv5 consists of three main parts. The backend is a standard convolutional backend as we know it from other neural networks (VGG, ResNet, DarkNet, etc). The backend extracts the feature maps from the input image and performs the geometrical pattern detections. As we go deeper through the backend, the extracted feature maps' resolution is decreasing, and the neural network detects larger and more complex geometrical shapes.

\begin{figure}[H]
    \includegraphics[width=12cm]{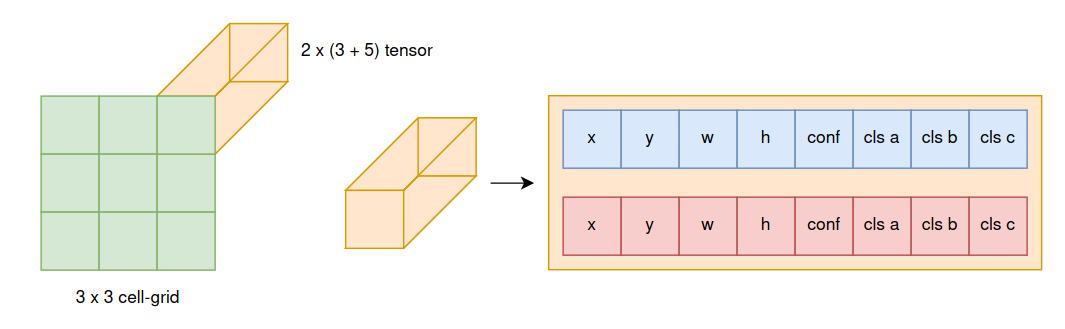}
    \caption{Image shows the simple example of the single output tensor that represents the 3x3 cell-grid where each cell predicts two bounding boxes of given X and Y position, width W and height H, all relative to the cell area and position, the bounding box detection confidence and three probabilities of object affiliation to the given classes.}
    \label{fig:yolo_ouptut_tensor}
\end{figure}

The second part is the "neck" stage. It takes the feature maps from several backend levels (different backend levels detect objects of different sizes and geometrical complexities) and combines them into three output scales.

The last part is the prediction park that uses 1x1 convolution layers to map different scales of concatenated feature maps from the "neck" stage into three output tensors.

For the 640x640x3 input image our network provides tensors of a 3x20x20x(N+5), 3x40x40x(N+5) and 3x80x80x(N+5) dimensions on the output. The 3 represents the number of anchor boxes predicted per cell, the 20x20 (40x40 and 80x80) gives the dimensions of the cell-grids that represent the 2D input signal, the N is the number of predicted classes (1 or 2 in our case).

\subsubsection{SSD}

The Single Shot Multibox Detector (SSD) is a fast object detector, originally presented in \cite{liu2016ssd}. The architecture consists of a feature extracting base-net (originally VGG16) and classifying layers. The base-net findes feature map using convolutional layers, with the number of features rising as we go deeper into the base-net. The classifying layers are connected to the base-net on several levels for different level features. For this experiment, we used the open source implementation available from \cite{pytorch-ssd}.


The SSD detector processes images or batch of images through net and compute output. The output is represented by the predicted object's location bounding box and confidence. For this output the architecture generates tens of thousands of priors, which are base bounding boxes similar to anchor boxes in Faster RCNN \cite{ren2016faster} from which we compute regression loss.

This architecture creates the base net and adds several predictors for finding different scales of objects. As we go deeper into the base-net, we assume finding a bigger object as the feature map corresponds to a bigger receptive field. The first predictor finds boxes after 14 convolution layers, therefore fails to find very small objects (feature map 38x38px).

From input size 300x300x3, the network return tensors 4x38x38x(N+4), 6x19x19x(N+4), 6x10x10x(N+4), 6x5x5x(N+4), 4x3x3x(N+4), 4x1x1x(N+4) output. The 4 or 6on the beginning represents the number of anchor boxes predicted per cell, the 38x38 (19x19, 10x10, 5x5, 3x3, 1x1) gives the dimensions of the cell-grids that represent the 2D input signal, the N is the number of predicted classes (1 or 2 in our case) and the 4 in the brackets refer to 4 coordinates for regression box size (relative x-min,y-min,x-max,y-max).

In our experiment, we use two base-nets. The first one is the original VGG16 and the second in MobileNet v2. The main difference is in the network's complexity (VGG16 has approx 138 mil. parameters and MobileNet v2 has approx. 2.3 mil.).

\subsubsection{Deep SVDD}

The Deep Support  Vector Data Description (Deep SVDD) is a convolutional neural network based anomaly detector (AD) presented in the \cite{svdd}. It is designed for the detection of anomalous samples (one class classification) and it already brought us good results on homogenous datasets.


This technique searches for a neural network transform, which maps the majority of the input data to a hypersphere with a center c and a radius R with a minimal volume. The samples, which fall into this area are considered as normal and the samples out of sphere as anomalous. The center of the hypersphere is computed from the mean features of the all input samples and the decision boundary is set by a threshold. The input data might be processed by an autoencoder. For our experiment, we used the implementation available from \cite{DeepSVDDimplementation}.

We used this model in our experiment even though it is not an object detector, because we wanted to prove whether  this architecture yields good results in the infected bee detection. As we mentioned in the introduction, the V.-mite infested bees are often deformed and the parasite might not be always visible, or present on the bee’s body. With the AD technique based approach, we could be able to detect those cases, or even bees with other problems.

For this experiment, we used 200 samples from the dataset \cite{schurischuster_stefan_2020_4085043}, because this method requires similar looking samples, which could not be sourced from our part of the dataset described above.

\subsubsection{Hyperparameters}

We trained all networks on the Nvidia GTX 2080 GPU card. The YOLOv5 and Deep SVDD were trained on a 640 by 640px images, the SSD was trained on 300 by 300px as in the original paper. The batch size was four images. We used the ADAM optimizer. For the YOLOv5 and SSD networks, we used the implementation’s default pretrained weights to boost the learning process and save computational resources. In all cases, we left all network weights unlocked for training.

We trained each YOLO model for 100 epochs and the performance of the models saturated after 30 epochs. The SSD models were trained again for 150 epochs and they saturated after 80 epoch. For the purpose of preventing overfitting on the training dataset, the model was chosen. The probability threshold was set to 0.4 for YOLO and 0.3 for SSD.

We let the defaults anchor boxes for the YOLOv5 model: [10,13, 16,30, 33,23] (P3/8), [30,61, 62,45, 59,119] (P4/16) and [116,90, 156,198, 373,326] (P5/32), as defined in \cite{jocher2020yolov5}. For the SSD model, the anchor boxes required more tuning and we set them for both base nets as follows:

\begin{specialtable}[H]
\scriptsize
\centering

\begin{tabular}{|c|c|c|c|c|}
\hline
Base Net & Feature Map Size & Shringkage & Anchor Box & Aspect Ratio \\ \hline \hline
\multirow{6}{*}{VGG 16} & 38 & 16 & (15, 30) & {[}1, 2{]} \\ \cline{2-5} 
 & 19 & 32 & (30, 60) & {[}1, 2{]} \\ \cline{2-5} 
 & 10 & 64 & (60, 105) & {[}1, 2{]} \\ \cline{2-5} 
 & 5 & 100 & (105, 150) & {[}1, 2{]} \\ \cline{2-5} 
 & 3 & 150 & (150, 195) & {[}1, 2{]} \\ \cline{2-5} 
 & 1 & 300 & (195, 240) & {[}1, 2{]} \\ \hline \hline
\multirow{6}{*}{MobileNet v2} & 19 & 16 & (15, 30) & {[}1, 2{]} \\ \cline{2-5} 
 & 10 & 32 & (30, 60) & {[}1, 2{]} \\ \cline{2-5} 
 & 5 & 64 & (60, 105) & {[}1, 2{]} \\ \cline{2-5} 
 & 3 & 100 & (105, 150) & {[}1, 2{]} \\ \cline{2-5} 
 & 2 & 150 & (150, 195) & {[}1, 2{]} \\ \cline{2-5} 
 & 1 & 300 & (195, 240) & {[}1, 2{]} \\ \hline
\end{tabular}

\caption{Setting of the SSD anchor boxes.}
\label{tab:ssd_anchor}
\end{specialtable}

The size of the SSD anchor boxes was set according to the size of the V-mites in our dataset, which varied in range between 15-25 px.

\subsubsection{Used Metrics}

To estimate the performance of the artificial intelligence models, we need to define metrics that will give us an idea of how well the model could solve a given problem after the training process.

One of the most common ways to express object detection capability to perform well is the mean average precision (mAP) metrics first introduced by the \cite{everingham2010pascal} as the mAP[0.5]. This metric denotes the relative number of object detection precision using the minimal intersection over union (IoU) with a value equal to or bigger than 0.5.

Another variant of the mAP metrics is the mAP[0.5:0.95], first used by \cite{lin2014microsoft}. In addition to mAP[0.5], mAP[0.5:0.95] calculates the average for all mAP values with the IoU level 0.5 up to 0.95 with 0.05 step.

\begin{figure}[H]
    \centering
    \includegraphics[width=6cm]{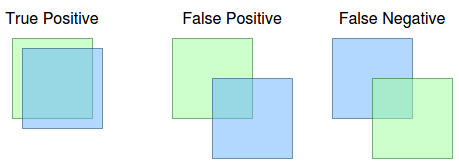}
    \caption{True positive, false positive and false negative results.}
    \label{fig:tpfp}
\end{figure}

The meaning of true positive, as well as false positive and false negative in context of the IoU metrics can be seen in figure \ref{fig:tpfp}. As can be seen in the figure, the result is a true positive if the ground truth (blue bounding box) and the detection (green bounding box) has a level of IoU at least 0.5. We obtain the false positive result if the neural network declares the detection, and there is no ground truth bounding box with at least 0.5 IoU. The false negative result means that the ground truth bounding box has no detection to match with at least 0.5 IoU. Note: the x value defines0.5 IoU from mAP[x] metric.

Then the precision and recall are given by:

\begin{equation}
precision = \frac{True Positive}{True Positive + False Positive} \\
recall = \frac{True Positive}{True Positive + False Negative}
\label{eq:precision_recall}
\end{equation}

Given something, we define the usual statistical metrics called the F1 score as:

\begin{equation}
F1 = 2 * \frac{precision * recall}{precision + recall}
\label{eq:f1_eq}
\end{equation}

\section{Results}

\subsection{YOLOv5}

The table \ref{tab:yoloresults} presents all performance numbers reached by YOLOv5 models trained on our dataset. All six models are always the "S '' and "X" variant of the YOLOv5, trained on three differently annotated modifications of the original data - the bees and V.-mite detection, the healthy and ill bees detection and the V.-mite only detection.

The primary purpose of our work is to track the health of the beehive population. From this perspective, the best results are shown by the X model trained on the Healthy and Ill bees dataset. It has the highest F1 score that expresses the relation between the precision and recall performance when detecting bees influenced by the V.-mite, but it also gives the best results from the mAP point of view where it outperforms other models trained on the same images but with different annotation.

\clearpage

\end{paracol}
\begin{figure}[H]
    \centering
    \includegraphics[width=18cm]{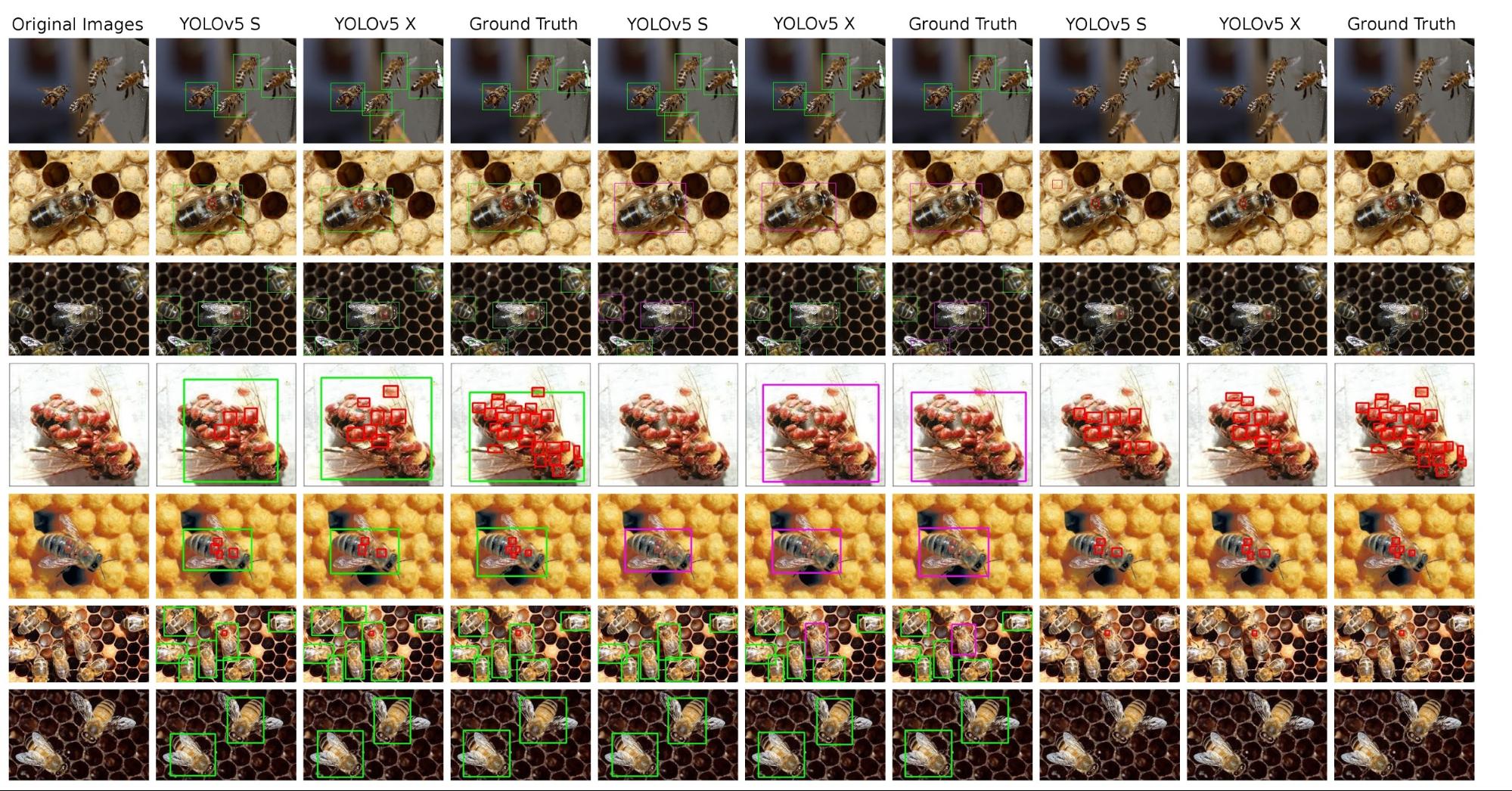}
    \caption{An overview of the detections performed on test images by the YOLO models, complemented by ground truth annotated images. 1st column: original image; 2nd - 4th column - model trained in bees and V.-mites detection; 5th - 7th column: models trained in healthy and ill bees detection; 8th - 10th column: models trained in V.-mite detection only.}
    \label{fig:yolo_results}
\end{figure}
\begin{paracol}{2}
\switchcolumn

\begin{specialtable}[H]
\scriptsize
\centering

\begin{tabular}{|c|c|c|c|c|c|c|c|c|c|c|}
\hline
\multicolumn{11}{|c|}{Bees and Varroa Mites Annotated Dataset} \\ \hline \hline
Metrics & \multicolumn{2}{c|}{mAP{[}0.5{]}} & \multicolumn{2}{c|}{mAP{[}0.5:0.95{]}} & \multicolumn{2}{c|}{F1} & \multicolumn{2}{c|}{Precision} & \multicolumn{2}{c|}{Recall} \\ \hline
Yolov5 model & S & X & S & X & S & X & S & X & S & X \\ \hline
Bees & 0.953 & 0.946 & 0.610 & 0.647 & 0.827 & 0.859 & 0.711 & 0.762 & 0.989 & 0.985 \\ \hline
V.-Mites & 0.649 & 0.726 & 0.260 & 0.276 & 0.638 & 0.650 & 0.657 & 0.609 & 0.620 & 0.696 \\ \hline
Average & 0.801 & \textbf{0.845} & 0.435 & \textbf{0.462} & 0.733 & \textbf{0.754} & 0.684 & \textbf{0.686} & 0.805 & \textbf{0.841} \\ \hline \hline
\multicolumn{11}{|c|}{Healthy and Ill Bees Annotated Dataset} \\ \hline \hline
Metrics & \multicolumn{2}{c|}{mAP{[}0.5{]}} & \multicolumn{2}{c|}{mAP{[}0.5:0.95{]}} & \multicolumn{2}{c|}{F1} & \multicolumn{2}{c|}{Precision} & \multicolumn{2}{c|}{Recall} \\ \hline
Yolov5 model & S & X & S & X & S & X & S & X & S & X \\ \hline
Healthy Bees & 0.932 & 0.938 & 0.589 & 0.676 & 0.826 & 0.865 & 0.722 & 0.824 & 0.964 & 0.910 \\ \hline
Ill Bees & 0.902 & 0.908 & 0.726 & 0.746 & 0.838 & 0.874 & 0.825 & 0.902 & 0.852 & 0.848 \\ \hline
Average & 0.917 & \textbf{0.923} & 0.658 & \textbf{0.711} & 0.832 & \textbf{0.870} & 0.774 & \textbf{0.863} & \textbf{0.908} & 0.879 \\ \hline \hline
\multicolumn{11}{|c|}{Varroa Mites Annotated Dataset} \\ \hline \hline
Metrics & \multicolumn{2}{c|}{mAP{[}0.5{]}} & \multicolumn{2}{c|}{mAP{[}0.5:0.95{]}} & \multicolumn{2}{c|}{F1} & \multicolumn{2}{c|}{Precision} & \multicolumn{2}{c|}{Recall} \\ \hline
Yolov5 model & S & X & S & X & S & X & S & X & S & X \\ \hline
V.-Mites & \textbf{0.777} & 0.752 & \textbf{0.328} & 0.327 & 0.656 & \textbf{0.666} & 0.585 & \textbf{0.626} & 0.746 & \textbf{0.712} \\ \hline
\end{tabular}
\caption{Test score achieved by the YOLOv5 models on all datasets variants. The best results in V.-mite detection achieved by the YOLOv5 X model for the Healthy and Ill bees dataset.}
\label{tab:yoloresults}
\end{specialtable}

Originally, we expected that the recognition of the tiny difference between the healthy bee and the one infected by the V.-mite would be the most challenging problem for the neural network. In the end however, the opposite was proven to be true.

\subsection{SSD}

SSD object detector results are presented in the table \ref{tab:ssdresults}. Just as in the previous experiment, we tested three datasets and two base nets. The shortcut mb2 stands for the MobileNetV2 and the shortcut vgg stands for the VGG16 architecture.

\clearpage

\end{paracol}
\begin{figure}[H]
    \centering
    \includegraphics[width=18cm]{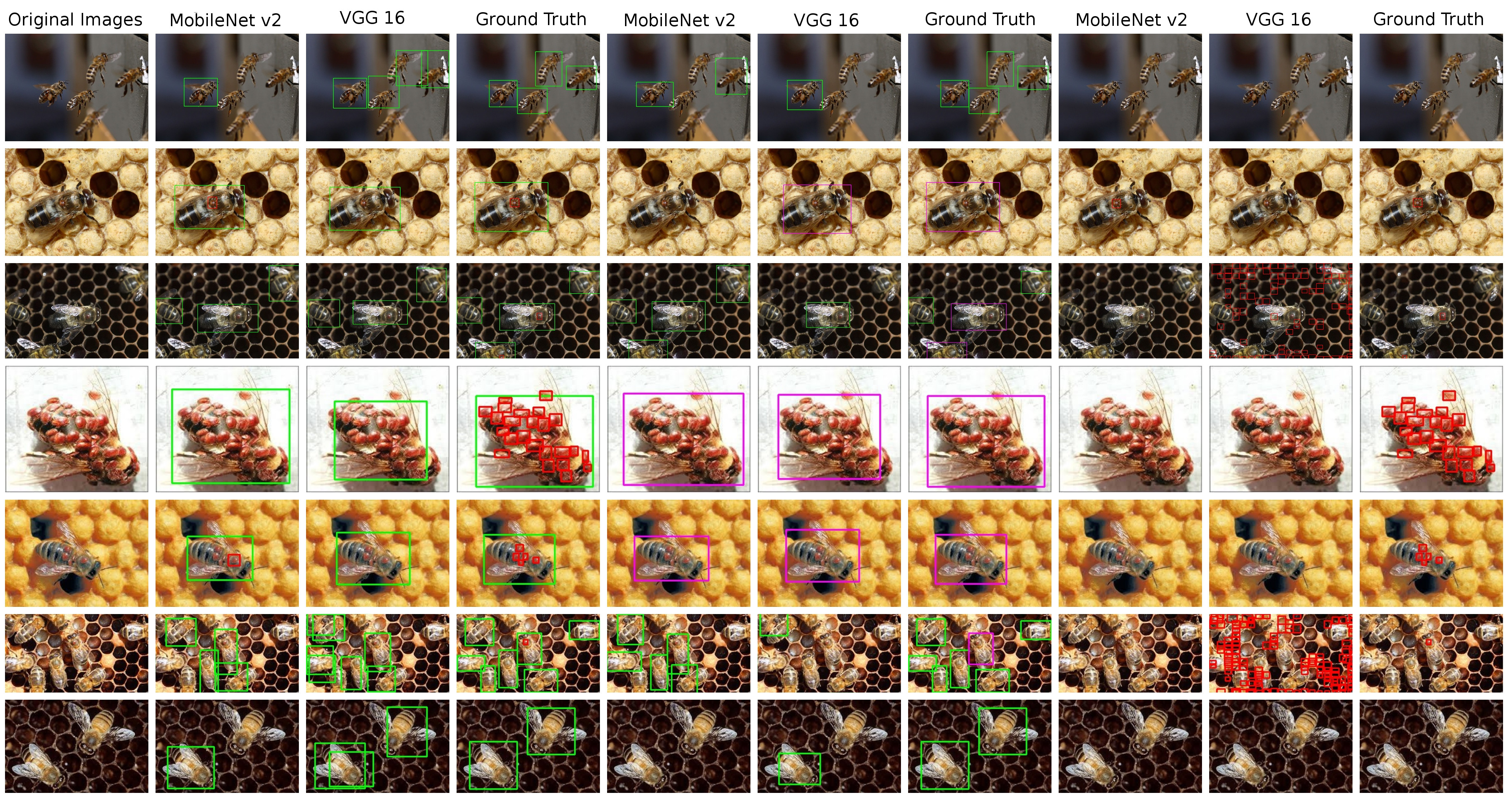}
    \caption{An overview of the detections performed on test images by the SSD models, complemented by ground truth annotated images. 1st column: original image; 2nd - 4th column - model trained in bees and V.-mites detection; 5th - 7th column: models trained in healthy and ill bees detection; 8th - 10th column: models trained in V.-mite detection only.}
    \label{fig:ssd_results}
\end{figure}
\begin{paracol}{2}
\switchcolumn

\begin{specialtable}[H]
\scriptsize
\centering

\begin{tabular}{|c|c|c|c|c|c|c|c|c|c|c|}
\hline
\multicolumn{11}{|c|}{Bees and Varroa Mites Annotated Dataset} \\ \hline \hline
Metrics & \multicolumn{2}{c|}{mAP{[}0.5{]}} & \multicolumn{2}{c|}{mAP{[}0.5:0.95{]}} & \multicolumn{2}{c|}{F1} & \multicolumn{2}{c|}{Precision} & \multicolumn{2}{c|}{Recall} \\ \hline
Base Net & VGG & MB2 & VGG & MB2 & VGG & MB2 & VGG & MB2 & VGG & MB2 \\ \hline
Bees & 0.564 & 0.621 & 0.393 & 0.464 & 0.537 & 0.748 & 0.475 & 0.855 & 0.617 & 0.665 \\ \hline
V.-Mites & 0.052 & 0.442 & 0.018 & 0.181 & 0.143 & 0.598 & 0.455 & 0.929 & 0.085 & 0.441 \\ \hline
Average & 0.308 & 0.532 & 0.206 & 0.323 & 0.340 & 0.673 & 0.465 & 0.892 & 0.351 & 0.553 \\ \hline \hline
\multicolumn{11}{|c|}{Healthy and Ill Bees Annotated Dataset} \\ \hline \hline
Metrics & \multicolumn{2}{c|}{mAP{[}0.5{]}} & \multicolumn{2}{c|}{mAP{[}0.5:0.95{]}} & \multicolumn{2}{c|}{F1} & \multicolumn{2}{c|}{Precision} & \multicolumn{2}{c|}{Recall} \\ \hline
Base Net & VGG & MB2 & VGG & MB2 & VGG & MB2 & VGG & MB2 & VGG & MB2 \\ \hline
Healthy Bees & 0.259 & 0.463 & 0.186 & 0.298 & 0.387 & 0.594 & 0.625 & 0.658 & 0.281 & 0.541 \\ \hline
Ill Bees & 0.587 & 0.502 & 0.504 & 0.478 & 0.680 & 0.659 & 0.733 & 0.769 & 0.635 & 0.577 \\ \hline
Average & 0.423 & 0.483 & 0.345 & 0.388 & 0.534 & 0.627 & 0.679 & 0.713 & 0.458 & 0.559 \\ \hline \hline
\multicolumn{11}{|c|}{Varroa Mites Annotated Dataset} \\ \hline \hline
Metrics & \multicolumn{2}{c|}{mAP{[}0.5{]}} & \multicolumn{2}{c|}{mAP{[}0.5:0.95{]}} & \multicolumn{2}{c|}{F1} & \multicolumn{2}{c|}{Precision} & \multicolumn{2}{c|}{Recall} \\ \hline
Base Net & VGG & MB2 & VGG & MB2 & VGG & MB2 & VGG & MB2 & VGG & MB2 \\ \hline
V.-Mites & 0.113 & 0.612 & 0.028 & 0.244 & 0.009 & 0.727 & 0.004 & 0.900 & 0.356 & 0.610 \\ \hline
\end{tabular}
\caption{Test score achieved by the SSD models on all datasets variants. The best results in V.-mite detection achieved by the SSD X model for the Healthy and Ill bees dataset.}
\label{tab:ssdresults}
\end{specialtable}

SSD performs worse than YOLOv5 in almost all metrics (with the exception of the F1 score on the V.-mites dataset). However the results show the potential of this detector for our task, when better results might be achieved by fine tuning and a further experimentation with the detector’s anchor boxes. As the MobileNetV2 is more recent and higher performance architecture, it clearly outperforms the VGG16 base net in all categories, especially on the V.-mites dataset.

\subsection{Deep SVDD}

Beside the YOLOv5 and SSD object detectors, we tried to use the Deep SVDD anomaly detector \cite{svdd} on the part of the dataset \cite{schurischuster_stefan_2020_4085043}. Using this approach, we expected a better recognition of the bees deformated by the V.-mite, but this method didn’t bring any satisfactory results. Possible explanations of the failure of this method could be, that the image features are too close to each other, or that the anomalous bees create a cluster inside the correct features.
\section{Discussion}

To compare the performance of both object detectors, we decided to use mAP[0.5] and F1 scores. The mAP[0.5] is a commonly used standard of object detector’s performance measurement. We selected the mAP[0.5] over the the mAP[0.5:0.95] as the latter metric is highly influenced by the size of detected objects and for smaller objects (V.-mite) it gives a significantly worse score than for large objects (bees). As the second metric, we chose F1 score as an expression of the relation between the precision and the recall, as we need both of these parameters to have high values to avoid false positives and false negatives detections.

The outcomes shown above prove that the object detectors can be used in the visual honey bee health state inspection and that our results are comparable with the other related research. The best results were obtained with the YOLOv5 neural networks trained on the Healthy and Ill bees dataset. In the S variant, YOLOv5 reaches 0.902 mAP[0.5] and 0.838 F1 score. For the X variant it reaches 0.908 mAP[0.5] and 0.874 F1 score. An overview of our results is shown in table \ref{tab:finalresults}.

\end{paracol}
\begin{specialtable}[H]
\scriptsize
\centering
\begin{tabular}{|c|c|c|c|c|c|c|c|c|}
\hline
\multicolumn{9}{|c|}{Detecting V.-Mite Presence} \\ \hline
 & \multicolumn{4}{c|}{mAP{[}0.5{]}} & \multicolumn{4}{c|}{F1 Score} \\ \hline
Training Dataset & YOLOv5 S & YOLOv5 X & SSD-VGG & SSD-MobNv2 & YOLOv5 S & YOLOv5 X & SSD-VGG & SSD-MobNv2 \\ \hline
\begin{tabular}[c]{@{}c@{}}Bees and V.Mites\\ (V.-mite score only)\end{tabular} & 0.649 & 0.726 & 0.052 & 0.442 & 0.638 & 0.650 & 0.143 & 0.598 \\ \hline
\begin{tabular}[c]{@{}c@{}}Health and Ill Bees\\ (ill bees score only)\end{tabular} & 0.902 & 0.908 & 0.587 & 0.502 & 0.838 & 0.874 & 0.680 & 0.659 \\ \hline
V.-Mites Only & 0.777 & 0.752 & 0.113 & 0.612 & 0.656 & 0.666 & 0.009 & 0.727 \\ \hline
\end{tabular}
\caption{Table shows the mAP[0.5] score at V.-mites detection (V.-mites or an infected bee) for all neural network models that we used in this work.}
\label{tab:finalresults}
\end{specialtable}
\begin{paracol}{2}
\switchcolumn

For the YOLOv5, the F1 score on the Bees and V.-mites dataset is lower than the comparable scores 0.95 and 0.99 achieved in \cite{schurischuster2020image} and \cite{bjerge2019computer}. We explain this difference with a higher difficulty of the object detection task in comparison with the object classification and with the use of a more complex dataset than in the papers above. Nevertheless, the resulting F1 score on the Healthy and Ill bees dataset is comparable to the mentioned papers. By our opinion, that distinguishing between those two classes can prove even more robust than the single V.-mite detection, because the infection is often connected with the bee body deformations and the V.-mite might not always be present on the bee’s body.

The SSD detector required more tuning than the YOLOv5, but it also gives promising results. The MobileNetV2 outperforms the VGG16 base net and brings the best F1 score (0.727) on the V.-mite dataset.

\section{Conclusion}

Our paper presents another approach with an online measurement potential to identify V.-mite infection in the beehive by using object detectors. To the best of our knowledge, there is no other paper that deals with the Varroosis detection problem this way, as most of the papers are studying only image classification. Moreover, we split our work into three directions by modifying the datasets to train three slightly different problems. The first variant was to train neural networks to detect both bees and V.-mite. The second approach was to detect healthy and V.-mite infested bees, and the third approach was to detect V.-mite only.

We tested three architectures, the YOLOv5 neural network,  the SSD neural network (both object detectors) and the SVDD, neural network-based anomaly detector. The YOLOv5 and SSD object detectors proved to be suitable for this task, however, the SVDD anomaly detector was not able to model the problem. We plan to use both YOLOv5 and SSD detectors in our future experiments.

Unlike the papers mentioned above, the YOLOv5 and SSD inference times should allow us to process the inspection results online. We plan to use this ability in our future work, where we would like to develop a monitoring system similar to the \cite{bjerge2019computer} with the emphasis on low cost and it being a more portable solution. We also plan to extend our dataset in order to monitor other bee infections and bee colony activities.

\vspace{6pt} 

\authorcontributions{Conceptualization, S.B; methodology, S.B., A.L and L.K.; software, A.L. and L.K.; validation, T.Z., M.H.; formal analysis, K.H., L.Z.; data curation, S.B.; writing---original draft preparation, S.B., A.L. and L.K.; writing---review and editing, T.Z., O.B., M.H.; visualization, A.L.; supervision, K.H. and L.Z.; All authors have read and agreed to the published version of the manuscript.}

\funding{The completion of this paper was made possible by the grant No. FEKT-S-20-6205 - "Research in Automation, Cybernetics and Artificial Intelligence within Industry 4.0" financially supported by the Internal science fund of Brno University of Technology.}

\institutionalreview{Not applicable}

\informedconsent{Not applicable}

\conflictsofinterest{The authors declare no conflict of interest. The funders had no role in the design of the study; in the collection, analyses, or interpretation of data; in the writing of the manuscript, or in the decision to publish the~results.} 

\end{paracol}
\reftitle{References}

\externalbibliography{yes}
\bibliography{texts/main.bib}

\end{document}